%% file: main_arxiv.tex
\documentclass{article}

\usepackage{microtype}
\usepackage{graphicx}
\usepackage{subfigure}
\usepackage{booktabs} 
\usepackage{multirow}
\usepackage{diagbox} 
\usepackage{bm}
\usepackage{graphicx}
\graphicspath{{Figures/}}  

\usepackage{hyperref}

\usepackage[accepted]{icml2024}

\usepackage{amsmath}
\usepackage{amssymb}
\usepackage{mathtools}
\usepackage{amsthm}

\usepackage[capitalize,noabbrev]{cleveref}

\theoremstyle{plain}

\theoremstyle{definition}

\theoremstyle{remark}

\input{abbrv.tex}

\usepackage[textsize=tiny]{todonotes}

\icmltitlerunning{OT-Transformers}

\begin{document}

\twocolumn[
\icmltitle{OT-Transformer: A Continuous-time Transformer Architecture with Optimal Transport Regularization}

\icmlsetsymbol{equal}{*}

\begin{icmlauthorlist}
\icmlauthor{Kelvin Kan}{equal,yyy}
\icmlauthor{Xingjian Li}{equal,comp}
\icmlauthor{Stanley Osher}{yyy}
\end{icmlauthorlist}

\icmlaffiliation{yyy}{Department of Mathematics, University of California, Los Angeles, USA}
\icmlaffiliation{comp}{Oden Institute for Computational Engineering and Sciences, University of Texas, Austin, USA}

\icmlcorrespondingauthor{Kelvin Kan}{kelvin.kan@math.ucla.edu}

\icmlkeywords{Transformers, Optimal Transport Theory, Neural ODE}

\vskip 0.3in
]

\printAffiliationsAndNotice{\icmlEqualContribution} 

\begin{abstract}
Transformers have achieved state-of-the-art performance in numerous tasks. In this paper, we propose a continuous-time formulation of transformers. Specifically, we consider a dynamical system whose governing equation is parametrized by transformer blocks. We leverage optimal transport theory to regularize the training problem, which enhances stability in training and improves generalization of the resulting model. Moreover, we demonstrate in theory that this regularization is necessary as it promotes uniqueness and regularity of solutions. Our model is flexible in that almost any existing transformer architectures can be adopted to construct the dynamical system with only slight modifications to the existing code. We perform extensive numerical experiments on tasks motivated by natural language processing, image classification, and point cloud classification. Our experimental results show that the proposed method improves the performance of its discrete counterpart and outperforms relevant comparing models.
\end{abstract}

\section{Introduction}
\label{sec:intro}

Transformers were first introduced in~\cite{vaswani2017attention} for natural language processing (NLP) tasks. The key feature of the model is the self-attention mechanism, which can capture dependencies of long sequences of data in a parallel manner. This renders the training of transformers more efficient than other architectures, such as RNNs and CNNs, especially when long sequences of data are involved. Since then, not only did it achieve state-of-the-art results in NLP~\cite{radford2019language}, but it also found various successful applications, including computer vision~\cite{dosovitskiy2021an}, program synthesis~\cite{chen2021evaluating}, computational biology~\cite{jumper2021highly}, speech processing~\cite{baevski2020wav2vec}, reinforcement learning~\cite{chen2021decision,lin2024transformersdecisionmakersprovable}, operator learning~\cite{li2023transformer,yang2023context} and climate modeling~\cite{gao2023earthformerexploringspacetimetransformers,nguyen2023climaxfoundationmodelweather,nguyen2024scalingtransformerneuralnetworks}.

The basic structure of a transformer architecture is transformer blocks, where self-attention is a key characteristic. In each transformer block, the self-attention layer can capture relationships within the input data in a parallel and efficient manner. The parallel computation of self-attention enhances the transformer's efficiency while preserving its representational power.

Each transformer block also incorporates a skip-connection structure. Inspired by the popular Neural ODE framework~\cite{chen2018neural}, we propose a continuous-time formulation for transformers, where the hidden states evolve over time according to an ODE. We further leverage optimal transport theory to regularize the hidden state dynamics. We justify this regularization both theoretically and experimentally.

\begin{figure}[h]
    \centering
    \hspace{-1em}
    \includegraphics[width=0.5\textwidth]{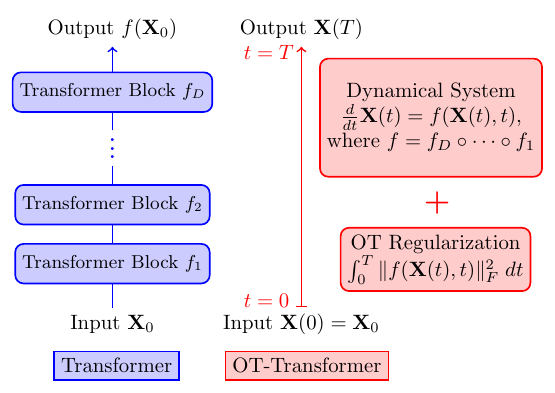}
    \caption{A schematic comparison between the transformer blocks of a \textbf{left:} vanilla transformer and \textbf{right:} OT-Transformer. OT-Transformer can directly reuse the pre-defined architecture $f_i$'s to parametrize the continuous-time dynamics, which only requires slight modification in the existing program.}
    \label{fig:diagram}
\end{figure}

We name our model OT-Transformer. Our approach is flexible and straightforward to implement in the sense that one can directly use a predefined transformer architecture to parametrize the ODE. This requires only slight modifications to existing code and opens up possibility for adapting existing architecture. When a single step forward Euler integration scheme is used, our model coincides with the original discrete transformer. Hence, OT-Transformer includes the original transformer as a special case. See~\Cref{fig:diagram} for a schematic comparison between the transformer blocks of a pre-defined transformer and OT-Transformer.

We summarize our contributions as follows:

\begin{itemize}
    \item We propose a continuous-time architecture of transformers. We composite transformer blocks to formulate an ODE governing the dynamics of the hidden states of the transformer. To the best of our knowledge, our approach is distinct from existing transformer models.

    \item Leveraging optimal transport theory, we use a regularization term penalizing the square arc length of the hidden state trajectory. We remark that the application of optimal transport to the design of transformer architecture remains underexplored and has shown very limited success.
    
    \item We demonstrate the effectiveness of the regularization. On the theoretical side, we apply optimal control theory to show that the unregularized training problem is ill-posed, that is, the solution is not unique and hence can be highly irregular. On the empirical side, our experimental results show that the regularization term improves generalization and leads to significantly more numerically stable training across different applications.
    \item Our experimental results show that our approach improves the performance of the vanilla architecture. In particular, it yields better performance with a reduced number of parameters, which leads to better memory efficiency at inference. In addition, our model outperforms existing continuous-time transformer models.
\end{itemize}

\section{Background}
In this section, we discuss the related work that motivate our approach.

\paragraph{Notations} In this paper, we use bold uppercase letter (e.g., $\bfX$) to denote matrices and bold lowercase letter (e.g., $\bfx$) to denote vectors. Moreover, we use $\bfx_j$ (resp. $\bfx_{i,j}$) to represent the $j$th column of $\bfX$ (resp. $\bfX_i$).

\paragraph{Transformers} 
In general, a transformer architecture is formulated as follows. Given an input $\bfZ=[\bfz_1,\bfz_2,...,\bfz_n] \in \mathbb{R}^{d_f \times n}$, where $n$ is the number of tokens and $d_f$ is their dimension, it first computes the input embedding of each token by
\begin{equation}\label{eq:input_embedding}
    \bfx_{0,j} = g_l(\bfz_{j}; \bfgamma_l), \quad \text{for} \quad j=1,2,...,n.
\end{equation}
Here $\bfx_{0,j} \in \mathbb{R}^{d}$. The input embedding $g_l$ parametrized by weights $\bfgamma_l$ embeds each token into a $d$-dimensional space and incorporates sequential order information into each token. Then, it is processed through a series of transformer blocks, where the output of each block serves as the input to the next. At each step, the model sequentially applies the operation $\bfX_{i+1}=f_{i+1}(\bfX_i)$ given by~\cite{thickstun2021transformer}\footnote{Layer normalization is commonly applied in each transformer block~\cite{xiong2020layer}. For brevity of exposition, it is omitted in the discussion. But it is included in our experiments.}
\begin{align}
     &\bfu_{i,j} = \bfx_{i,j} + \sum_{h=1}^H  \bfW_i^h \bfV_i^h \bfX_i \, \text{softmax} \left( \frac{(\bfK_i^h \bfX_i)^\top \bfQ_i^h \bfx_{i,j}}{\sqrt{k}} \right), \label{eq:self-attention}\\
     &\bfx_{i+1,j} = \bfu_{i,j} + g_{f}(\bfu_{i,j};\bftheta_{i}), \label{eq:fully-connected}
\end{align}
for $j=1,2,...,n$, and $i=0,1,...,D-1$, where $D$ is the total number of transformer blocks, $H$ is the number of self-attention heads, $\bfQ_i^h,\bfK_i^h, \bfV_i^h \in \mathbb{R}^{k \times d}$ are known as query, key, and value matrices, and $\bfW_i^h \in \mathbb{R}^{d \times k}$. In~\eqref{eq:fully-connected}, a fully connected layer $g_{f}$, parametrized by weights $\bftheta_{i}$, is applied individually to each of the $n$ tokens. The first equation~\eqref{eq:self-attention} is known as self-attention layers and is the key feature of transformer architectures. Their matrix multiplication formulation enables the parallel computation of dependencies among tokens, rendering them particularly effective for handling long sequences of tokens, that is, when $n$ is large. This self-attention mechanism enables models to focus on the most relevant parts of an input sequence, adapting dynamically to the context. Its flexibility allows it to capture complex, long-distance relationships within data, different from CNNs which primarily focus on local patterns, and RNNs, which experience a sharp performance decrease with long sequences. Such features make transformers particularly powerful for tasks such as language understanding and image recognition. This series of transformer blocks is also called an encoder in the literature.

Eventually, $\bfX_D$ is either passed to a decoder comprising another series of transformer blocks and then a multilayer perceptron (MLP) for sequence generation tasks or directly to an MLP for various downstream tasks, including classification and regression. 
The transformer output $\tilde{\bfy}$ is therefore computed by 
\begin{equation}\label{eq:output_layer}
    \tilde{\bfy} = g_o(\bfX_{D}; \bfgamma_o),
\end{equation}
where $g_o$ is either the composition of a decoder and an MLP or an MLP, parametrized by weights $\bfgamma_o$.

\paragraph{ResNets and Neural ODEs} Residual networks (ResNets)~\cite{he2016deep} are an extensively employed model which features a skip-connection structure in their layers. Given input $\bfx_0$, the output of the $i$th layer is computed by
\begin{equation}\label{eq:ResNet}
    \bfx_{i+1} = \bfx_{i} + g_{i}(\bfx_i).
\end{equation}
Here, $g_i$ is a network layer, and the skip-connection~\eqref{eq:ResNet} is a key feature of ResNet. This architecture is often compared with the explicit Euler discretization of an ordinary differential equation (ODE)~\cite{weinan2017proposal,haber2017stable,ruthotto2020deep}. Based on this insight,~\cite{chen2018neural} proposed Neural ODEs (NODEs), whose formulation is given by
\begin{equation}\label{eq:NODE}
    \frac{d\bfx(t)}{dt} = f_{\text{NODE}}(\bfx(t), t).
\end{equation}
Here $t \in [0, T]$ is artificial time and $f_{\text{NODE}}$ is a neural network parametrizing the dynamics. Given an input $\bfx(0)$, the final output $\bfx(T)$ is obtained by integrating~\eqref{eq:NODE}. A notable and relevant advantage of Neural ODEs is their parameter efficiency, as the continuous formulation allows them to model complex transformations over time with fewer parameters compared to traditional architectures. 

\paragraph{OT-based CNFs} A prominent application of NODEs is continuous normalizing flows (CNFs)~\cite{chen2018neural}. CNFs use~\eqref{eq:NODE} to paramtrize invertible mappings between a standard Gaussian distribution and an unknown target distribution. The ill-posed nature of the CNF formulation can often add to the complexity and computational cost for solving a problem.
Optimal transport (OT) based regularization has prominent applications in CNFs and is a powerful tool in improving accuracy and at times reducing cost. Among the infinitely many mappings between the two distributions, OT-based CNFs ~\cite{finlay2020train,yang2020potential,onken2021ot,vidal2023taming} target to find the optimal transport mapping. This is done by incorporating into the training objective regularization term(s) enforcing straight trajectories in~\eqref{eq:NODE}. This renders the training problem well-posed~\cite{huang2023bridging,zhang2023mean}. The straight trajectories also offer numerical advantages, as they make the numerical integration of~\eqref{eq:NODE} more tractable.

\section{OT-Transformers}
In this section, we introduce the continuous-time transformer with optimal transport regularization (OT-Transformer). A key feature of OT-Transformer is that, the model uses a combination of transformer blocks and NODE formulation. Specifically, the model parametrizes an ODE using transformer blocks, with the embedded inputs~\eqref{eq:input_embedding} serving as the initial state of the ODE, and the terminal state will be passed to the output layer~\eqref{eq:output_layer}. An optimal transport regularization is used in the training problem, and we demonstrate its benefits empirically and theoretically.

\paragraph{Model Formulation} Motivated by the connection between ResNet and neural ODEs, and the inherent skip-connection structure of transformer blocks~\eqref{eq:self-attention} and~\eqref{eq:fully-connected}, we formulate a continuous-time transformer.

Given an input sequence $\bfZ=[\bfz_1,\bfz_2,...,\bfz_n]$ of length $n$, we first apply the input embedding~\eqref{eq:input_embedding} to obtain the initial state $\bfX(0)=\bfX_0 \in \mathbb{R}^{d \times n}$. The dynamics of the hidden state is then governed by the ODE\footnote{We found that including the time variable $t$ as an input yields similar performance to excluding it. Therefore, in our implementation, we do not include the time variable $t$. That is, the right-hand side of the ODE is $f(\bfX(t); \theta)$. This simpler option allows us to directly reuse pre-defined transformer block architecture.}
\begin{equation}\label{eq:cts_self_attention}
    \frac{d\bfX(t)}{dt} = f(\bfX(t), t; \bftheta), \quad \text{for} \quad t \in [0, T],
\end{equation}
where $f$ is the composition of a sequence of transformer blocks defined in~\eqref{eq:self-attention} and~\eqref{eq:fully-connected}, that is, $f=f_D \circ f_{D-1} \circ ... \circ f_1$, and $\bftheta$ collectively denotes their trainable parameters $\bftheta_i$, $\bfK^h_i$, $\bfV^h_i$, $\bfQ^h_i$ and $\bfW^h_i$ for all $h$ and $i$. In real implementation, we adopt a discretize-then-optimize approach~\cite{onken2020discretize,onken2021ot} and compute the terminal state $\bfX(T)$ by using numerical integration schemes such as forward Euler or Runge–Kutta methods~\cite{butcher2016numerical}. Finally, we obtain the transformer output $\tilde{\bfy}$ by applying~\eqref{eq:output_layer} to the terminal state $\bfX(T)$.

Our framework is flexible in that it can be applied to almost any existing transformer architectures. It can directly reuse the architecture of an existing transformer's input embedding, decoder and output layers and use its transformer blocks $f_i$'s to construct the ODE~\eqref{eq:cts_self_attention}\footnote{It is possible to construct a continuous-time formulation for the decoder, where the decoder is used to formulated a second ODE. In this work, we focus on the continuous-time formulation for the encoder only and leave that for future work.}. This only requires slight modifications to existing code. Our framework generalizes the discrete formulation of transformer blocks to continuous-time, effectively enables a continuous-depth formulation of transformer blocks. When $T=1$ and a single step forward Euler integration scheme is used, our framework is identical to the original discrete transformer formulation. Hence, our framework is consistent with the standard transformer architecture.

\paragraph{Problem Formulation} 
We formulate the training objective as
\begin{equation}\label{eq:training_obj}
    \min_{\bftheta, \bfgamma} \; \mathbb{E}
    \left\{ L(\bfX(T), \bfy; \bfgamma) + \frac{\lambda}{2dn} \int_0^T \| f(\bfX(t), t; \bftheta) \|_{F}^2 \; dt \right\}.
\end{equation}
Here, the expectation is taken over the input-output pairs $(\bfZ, \bfy)$, $\| \cdot \|_F$ denotes the Frobenius norm, $\bfgamma$ collectively denotes the weights of the input embedding $\bfgamma_l$ and output layer $\bfgamma_o$, $\bftheta$ collectively denotes the weights of the transformer blocks $\bfW^h_i$, $\bfQ^h_i$, $\bfK^h_i$, $\bfV^h_i$, and $\bftheta_i$'s. The loss function $L$ measures the difference between the target output $\bfy$ and model output $\tilde{\bfy}(\bfX(\bfT); \bfgamma)$. For instance, in classification~\cite{dosovitskiy2021an} and sequence generation~\cite{vaswani2017attention} tasks,
one commonly uses the softmax loss. The second term is a transport cost regularization penalizing the squared norm of the velocity (the right hand side of the ODE). It enhances the regularity of the hidden state dynamics~\eqref{eq:cts_self_attention} by promoting more constant speed and straighter state trajectories. In practice, this regularization term is computed easily as it is calculated alongside the numerical integration of the ODE~\eqref{eq:cts_self_attention}. The regularization term is normalized by $1/dn$, where $dn$ is the dimension of $f$. This normalization accounts for the size of $f$, ensuring the regularization term remains consistent across different dimensions. The regularization parameter $\lambda$ balances the effects of the two terms.

\paragraph{Empirical Benefits of Transport Cost} As we will demonstrate empirically in our experiments, the transport cost improves the model's effectiveness. The transport cost serves as a regularizer and can stabilize the training process; without it, the model is more prone to experiencing exploding or vanishing gradients. Moreover, the generalization (i.e., performance on unseen data) of the model is enhanced, thanks to the more regular hidden state dynamics. Interestingly, the regularized model can also achieve a lower data-fitting loss for the training data, despite the incorporation of the regularization term. This occurs because the optimization process is stochastic, with each model update based on a batch of data rather than the entire training set. The regularization term, however, encourages broader generalization across the entire dataset.

\paragraph{Theoretical Benefits of Transport Cost} We theoretically demonstrate the purpose of the transport cost. Specifically, using optimal control theory~\cite{kirk2004optimal, liberzon2011calculus}, we show that the training problem is ill-posed without the transport cost regularization. In particular, the solution is not unique and thus can be highly irregular. We build upon the anaylses of~\cite{zhang2023mean,gu2024combining}, which study OT-based CNFs for learning a marginal distribution. We modify their approach to adapt to the case where the solution is conditional on $\bfy$.

For a given target output $\bfy$, 
optimal control theory~\cite{fleming2012deterministic, liberzon2011calculus}
states that there exists a potential function $\Phi_{\bfy}: \mathbb{R}^{d \times n} \times [0,T] \to \mathbb{R}$, where the optimal $f$ for~\eqref{eq:training_obj} can be represented by 
\begin{equation}\label{eq:feedback}
    f(\bfX, t) = - \frac{dn}{\lambda} \nabla \Phi_{\bfy}(\bfX, t),
\end{equation}
where the gradient $\nabla \Phi_{\bfy}(\bfX, t)$ is taken with respect to the first argument $\bfX$.
This is analogous to classical physics, where $\bfX$ moves in a manner to minimize its potential. Optimal control theory further states that the Hamilton-Jacobi-Bellman (HJB) equation~\cite{bellman1954dynamic, evans2010partial} is an optimality condition characterizing the optimal value $\Phi_\bfy$ and is given by
\begin{align}
    \begin{split}
    -\partial_t \Phi_{\bfy}(\bfX, t) + H(\bfX, \nabla \Phi_\bfy(\bfX, t)) &= 0,  \\
    \Phi_\bfy(\bfX, T) &= L(\bfX, \bfy),\label{eq:HJB}
    \end{split}
\end{align}
where the Hamiltonian $H:\mathbb{R}^{d \times n} \times \mathbb{R}^{d \times n}  \to \mathbb{R}$ is given by
\begin{equation}\label{eq:Hamiltonian}
    H(\bfX, \bfP) = \sup_f -\langle \bfP, f(\bfX, t) \rangle - \frac{\lambda}{2 dn} \| f(\bfX, t)\|_{F}^2,
\end{equation}
with $\langle \cdot,\cdot \rangle$ representing the Frobenius inner product, $\bfP$ is the adjoint variable to the system and is introduced by the Pontryagin Maximum Principle~\cite{mangasarian1966sufficient, fleming1975deterministic}.
We see that when $\lambda=0$, i.e., when the transport cost is absent, the Hamiltonian cannot be defined properly and equals infinity. 
Therefore, there is no well-defined HJB equation, and the training problem~\eqref{eq:training_obj} becomes degenerate.
As such there are infinitely many choices of $f$ that minimize the data fidelity term in~\eqref{eq:training_obj}, including some highly irregular ones. For instance, $f$ can produce a zig-zagging hidden state trajectory or move to the target location instantly and then remain stationary. These irregular paths can pose challenges in numerical integration and result in numerical instability during training, as demonstrated by our experiments. On the other hand, the addition of the transport cost promotes the uniqueness and regularity of the solution. In short, the training problem is well-posed only if the corresponding HJB equation is well-posed~\cite{lasry2007mean,bensoussan2013mean}.

\section{Related Work}\label{sec:related} 

This section provides a review of relevant work.

\paragraph{Continuous-time Architecture} There has been some works on a continuous-time interpretation of transformers. And there is a key distinction between the formulations of OT-Transformer and existing models. In OT-Transformer, we use the composition of all transformer blocks to parametrize a single dynamical system~\eqref{eq:cts_self_attention} governing the hidden states. To the best of our knowledge, the existing works use each transformer block to parametrize a dynamical system. For a transformer with $D$ transformer blocks, the continuous-time model is represented as the output of $D$ different dynamical systems. In particular, it is formulated as
\begin{equation}
\label{eq:block_dynamics}
\begin{aligned}
    \bfX_0(0) &= \bfX_0,  \\
    \bfX_i(0) &= \bfX_{i-1}(T), &  \text{for} \; 1 \leq i \leq D-1, \\
    \frac{d \bfX_i(t)}{dt} &= \hat{f}_i(\bfX_i(t), t; \hat{\bftheta}_i), & \text{for} \; t \in [0, T], \; 0 \leq i \leq D-1, 
\end{aligned}
\end{equation}
where $\hat{f}_i$ is the $i$th transformer block parametrized by weights $\hat{\bftheta}_i$ and defined in~\eqref{eq:self-attention} and~\eqref{eq:fully-connected}, except that the fully-connected layer~\eqref{eq:fully-connected} has no skip-connection.

This formulation is introduced in~\cite{baier2020n}, the model of which is conceptually the closest to our approach. Here, we highlight several key differences between their work and ours. Firstly, they only conduct the simple task of determining the parity of a binary sequence, rather than investigating its performance in general applications. When their approach is applied, it fails to improve performance over the vanilla transformer and instead degrades it. While they also propose the use of the transport cost, the regularization cannot improve the performance of their model when the sequence length exceeds eight. We observe similar issues when testing their model on other applications; see~\Cref{sec:experiment}. This is potentially due to their choice of formulation. Specifically, in~\eqref{eq:block_dynamics}, as the model transitions from one transformer block to the next, it effectively switches to a different dynamical system, introducing non-smoothness to the overall dynamics. This undermines the purpose of the transport cost regularization, which seeks to obtain a continuous and more constant velocity. In contrast, our model is formulated using only one dynamical system. The resulting dynamics is smoother and thus inherently better suited to incorporate the transport cost regularization. This is evident in our experimental results, while the regularization can always improve the generalization of OT-Transformer to a significant extent, this is not the case with their model; in certain scenarios, the regularization may even degrade their model's performance. Moreover, they do not provide theoretical analysis to the regularization or demonstrate its numerical advantages. We also mention that, while~\cite{baier2020n} proposes alternative formulations for further investigation, it does not consider ours, highlighting the novelty and non-triviality of our approach.

Since then, there have been a number of follow-up works that build on the formulation~\cref{eq:block_dynamics} to perform different tasks, including sequence generation~\cite{lu2020understanding,li2021ode,li2022ode,zhong2022a}, time series forecasting~\cite{xu2023dynamic,cheng2024rktrans}, and image classification~\cite{niu2024efficient,okubo2024cost}. But most of these methods only use the formulation~\eqref{eq:block_dynamics} as motivation and are discrete architectures in nature, and none of them consider transport cost regularization in their approach. Moreover, these models focused on a specific type of application and not general-purpose.

In order to access the performance of our OT-Transformer more comprehensively, we also include the existing transformer formulation~\eqref{eq:block_dynamics} as a benchmark in our experiments. It is referred to as ``N-ODE Transformer" in our experimental results, following the terminology in~\cite{baier2020n}.

\paragraph{Mathematical Analysis} There have been works that theoretically analyze a continuous-time formulation of transformers. In~\cite{geshkovski2023mathematical,geshkovski2024emergence}, they show that a continuous-time formulation can be interpreted as an interacting particle system, where each token can be perceived as a particle. They demonstrate that there is a clustering behavior among the tokens. Since then, there has been a number of works that further investigate the dynamics of tokens through this interpretation, including~\cite{adu2024approximate,bruno2024emergence,biswal2024identification,geshkovski2024measure,karagodin2024clustering}, to name a few. However, we note that the aforementioned work is primarily theoretical and lacks evaluations beyond toy experiments. In~\cite{sander2022sinkformers}, they show that, under some restriction on the weights, a continuous-time formulation of self-attention layers can be interpreted as a gradient flow. However, no experiments have been conducted following this analysis.

\section{Experimental Results}\label{sec:experiment}
We demonstrate the advantage of our proposed OT-transformers through four extensive experiments arising from point cloud classification, image classification, and text sentiment analysis. 

For each task, we use commonly used transformer architectures as baselines. All the hyperparameters of the experiments, including architectures of baseline models, number of epochs, learning rates, layer normalization, etc., are identical to those used in~\cite{sander2022sinkformers}. We also compared against N-ODE Transformer, an existing continuous-time transformer formulation which is introduced in~\cite{baier2020n} and has been considered in other works. For details about the formulation and specific applications, see the discussion in~\Cref{sec:related}. In the reported results, we refer to N-ODE Transformer with and without transport cost as unregularized N-ODE Transformer and regularized N-ODE Transformer, respectively.

For the continuous-time models, we employ the same architectures as the baselines but with a reduced hidden dimensions or number of layers for the transformer blocks. This is for investigating their parameter efficiency. 
To demonstrate the effectiveness of the transport cost on OT-Transformer, we also perform the experiments with $\lambda=0$ in~\eqref{eq:training_obj}, effectively creating an unregularized model. We label this model unregularized OT-transformer in the reported results. For the continous-time models, we use an explicit Euler scheme to numerically integrate the dynamical systems.

For more details of the experiments, we refer our readers to~\cref{sec:exp_details}. Our program is implemented using PyTorch~\cite{paszke2017automatic} and executed using NVIDIA A100 GPUs.

\subsection{Point Cloud Classification}
We use the ModelNet 40 dataset~\cite{wu20153d}, which is among the most widely used benchmark for point cloud classification~\cite{uy2019revisiting}. The dataset contains roughly 10,000 Computer-Aided Design (CAD) models that are categorized into 40 distinct classes, including common objects such as airplanes, cars, and furniture.

We experiment with the Set Transformer model~\cite{lee2019set}. It has an encoder-decoder architecture and is specifically designed to process unordered data, such as point clouds, ensuring that the output remains permutation invariant to its input. Following the setup of~\cite{sander2022sinkformers}, we use the baseline architecture with two Induced Self Attention Blocks (ISABs)~\cite{lee2019set} in the encoder, where each ISAB contains two transformer blocks, and experiment with 5,000 uniformly sampled points for each shape. For the continuous-time models, we use the same architecture except that we put a fully-connected layer before the transformer blocks so that the dimension is consistent for continuous-time dynamics. Also the hidden dimensions $d$ and $k$ of the ISABs are reduced from $256$ to $200$. This reduces the number of parameters for the ISABs by $24\%$. 

\begin{figure}[h]
    \centering
    \includegraphics[width=0.48\textwidth]{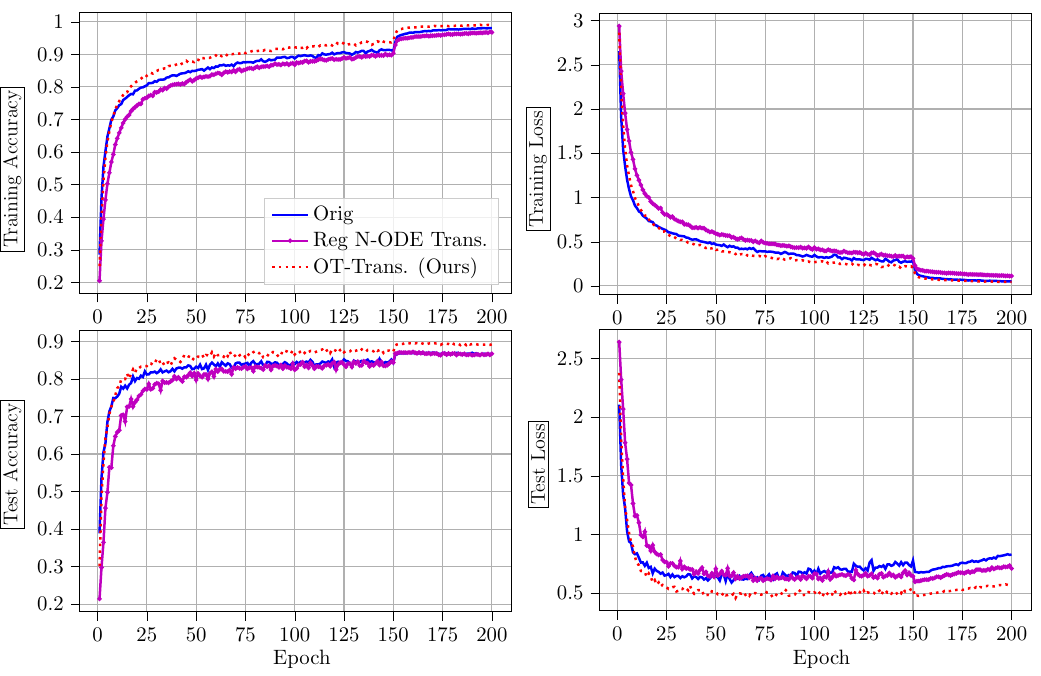}
    \vspace{-2em}
    \caption{Accuracy and data-fitting loss for the point cloud experiment (averaged over five trials)}
    \label{fig:pointcloud}
\end{figure}

We perform the experiment over five random trials and report the best test accuracies in~\Cref{tab:pointcloud}. The unregularized continuous-time models encountered gradient explosion, resulting in NaN outputs, and the issue persists even with slight regularization. We found that the models never suffered from gradient explosion with sufficient regularization, indicating that transport cost effectively stabilizes the training process. Hence, we only report the performance of the regularized models. The baseline Set Transformer obtains an average test accuracy of 87.4\%. The regularized N-ODE Transformer achieves an accuracy of 87.5\%, indicating negligible improvement over the vanilla model. Our OT-Transformer shows a sizable improvement and reports an average 89.9\% test accuracy even with a smaller model. From the learning curves in~\Cref{fig:pointcloud}, we see that our model reports a lower data-fitting loss for training data compared to the vanilla model, despite the inclusion of a regularization term,

\begin{table}[h]
    \centering
    \caption{Number of parameters for the transformer blocks, mean test accuracy and standard deviation (std) over five trials for the point cloud experiment.}
    \vspace{1em}
    \begin{tabular}{@{}lcc@{}}
        \toprule
        \textbf{Method/Exp.} & \textbf{Para. Count} & \textbf{Test Accuracy} \\ 
        \midrule
        Baseline & 0.86M & $87.4\% \pm 0.45\%$ \\ 
        Reg. N-ODE Trans. & 0.65M & $87.5\% \pm 0.51\%$  \\ 
        OT-Trans. (Ours) & 0.65M & $\mathbf{89.9} \textbf{\%} \bm{\pm}\mathbf{ 0.42} \textbf{\%}$ \\ 
        \bottomrule
    \end{tabular}
    \label{tab:pointcloud}
\end{table}

\subsection{Image Classification}
To further demonstrate the applicability of our proposed method, we also perform experiments on imaging tasks.
We consider the Vision Transformer (ViT), which was introduced in~\cite{dosovitskiy2021an}. Since then, the model and its variants have achieved state-of-the-art performance in computer vision tasks~\cite{ruan2022vision,xia2024vit}. The key feature of ViTs is that they divide an image into fixed-size patches, which are treated as sequences of data. ViTs then apply self-attention mechanisms to capture relationships between these patches, enabling it to learn complex structures across the entire image. We perform two image classification experiments following the same setup as in~\cite{sander2022sinkformers}.
\paragraph{MNIST Classification}
We first conduct a small-scale image classification experiment with the MNIST dataset~\cite{lecun1998mnist}. Following~\cite{sander2022sinkformers}, the baseline model ViT has one transformer block with a single-head self-attention layer and no fully-connected layer. Since it has only one transformer block, N-ODE Transformer and our OT-Transformer share the same formulation, and we report the results as OT-Transformer. 

\begin{figure}[h]
    \centering
    \includegraphics[width=0.48\textwidth]{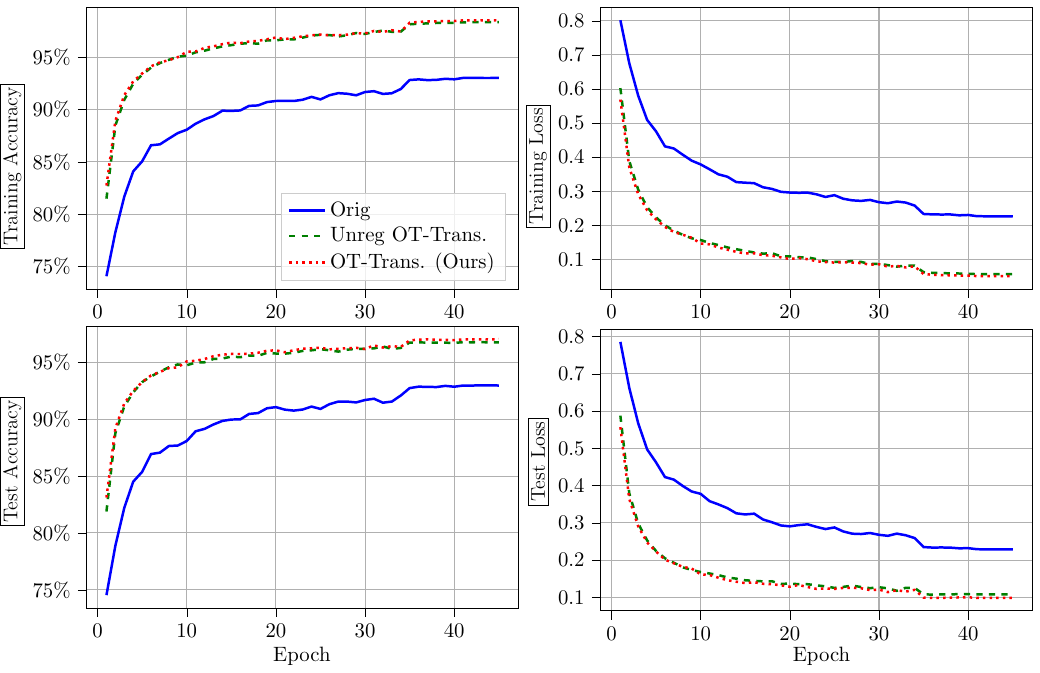}
    \vspace{-2em}
    \caption{Accuracy and data-fitting loss for the MNIST image classification experiment (averaged over five trials)}
    \label{fig:mnist}
\end{figure}

The OT-Transformer uses the same model architecture as the baseline model, except that the hidden dimensions $d$ and $k$ of the self-attention layer are reduced to 64 from 128. This reduces the number of parameters by over $80\%$. 
The experiments are conducted over five random trials. The best test accuracies are reported in~\Cref{tab:mnist}. OT-Transformer demonstrates significant improvements over the baseline in both accuracy and model efficiency. The baseline model achieved a test accuracy of 93.0\%. The unregularized OT-Transformer improves the test accuracy to 96.8\%, although it uses a much smaller model architecture. The transport cost regularization further improves the test accuracy to 97.1\% while maintaining the same reduced parameter count. Notably, OT-Transformer also exhibits significantly lower standard deviation across five trials when compared to the baseline and unregularized model, indicating enhanced stability and reliability in its performance. Interestingly, when we compare the learning curves of the unregularized and regularized OT-Transformers in~\Cref{fig:mnist}, we observe that including the transport cost regularization also reduces the training loss for data-fitting and accuracy. 

\begin{table}[h]
    \centering
    \caption{Number of parameters for different models, mean test accuracy and standard deviation over five trials for the MNIST image classification experiment.}
    \vspace{1em}
    \begin{tabular}{@{}lcc@{}}
        \toprule
        \textbf{Method/Exp.} & \textbf{Para. Count} & \textbf{Test Accuracy} \\ 
        \midrule
        Baseline & 93k & $93.0\% \pm 0.69\%$ \\ 
        Unreg. OT-Trans. & 18k & $96.8\% \pm 0.23\%$ \\ 
        OT-Trans. (Ours) & 18k & $\mathbf{97.1} \textbf{\%} \bm{\pm}
        \mathbf{0.16\%}$ \\ 
        \bottomrule
    \end{tabular}
    \label{tab:mnist}
\end{table}

\begin{figure}[h]
    \centering
    \includegraphics[width=0.48\textwidth]{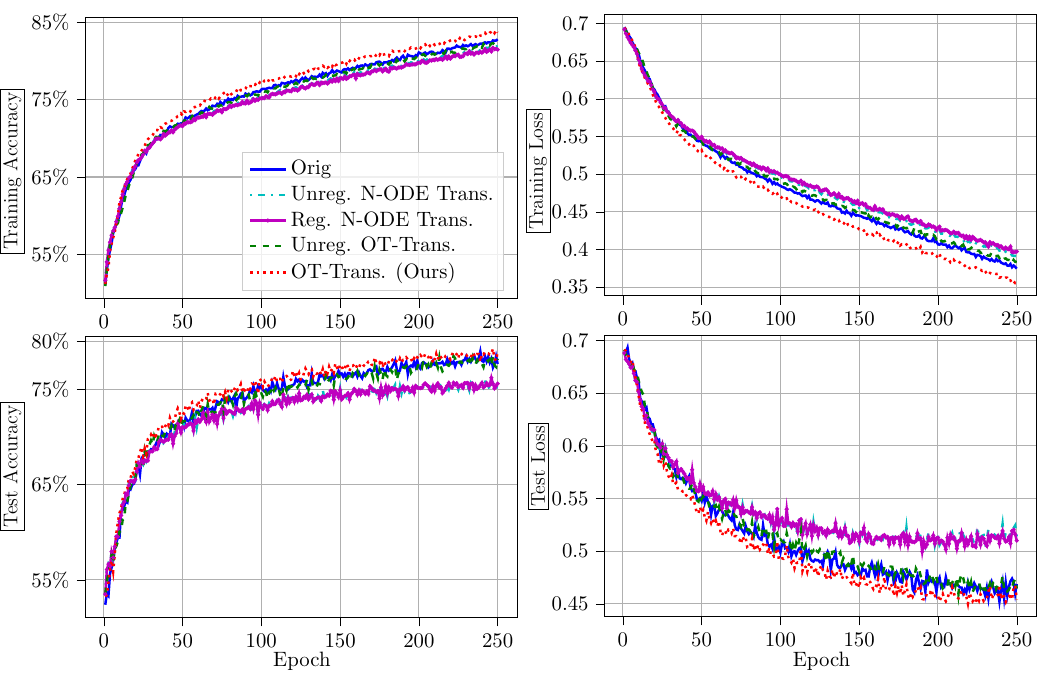}
    \vspace{-2em}
    \caption{Accuracy and data-fitting loss for the cats and dogs image classification experiment}
    \label{fig:catdog}
\end{figure}

\paragraph{Cats and Dogs Classification}
We perform experiments on a binary cats and dogs image classification task, following~\cite{sander2022sinkformers}. The baseline ViT has six layers of transformer blocks. We choose the continuous-time counterparts to have five layers; this reduces the number of parameters for the transformer blocks by around $20\%$. 
We report in~\Cref{tab:catdog} the test accuracies after the last epoch, which demonstrate a more significant improvement.
The best test accuracy is also reported in \cref{sec:exp_details}, where our model also performs best. We observe again that our OT-Transformer has the best performance and obtains a test accuracy of $79.0\%$, improving from the baseline's $77.6\%$. The standard deviation of the test accuracy, at $0.31\%$, is significantly lower than the baseline value of $0.86\%$, showing our proposed approach is more robust and reliable. We also observe that incorporating the transport cost regularization improves generalization and stability of OT-Transformer; without it, the average and standard deviation of test accuracy worsen to $78.2\%$ and $0.39\%$, respectively. Both the unregularized and regularized N-ODE Transformers report a test accuracy of 75.6\%, which is worse than the baseline model, making them undesirable methods for the problem. Unlike our model, incorporating the regularization also has little effect on the performance of N-ODE Transformer. This is likely due to the incompatibility of N-ODE Transformer and the regularization; see~\Cref{sec:related}.  We report the learning curves in~\Cref{fig:catdog}. When we compare the learning curves of the unregularized and regularized OT-Transformers, we see that including the transport cost regularization also improves the training loss for data-fitting and accuracy.

\begin{table}[h]
    \centering
    \caption{Number of parameters for the transformer blocks, mean test accuracy after the last epoch and standard deviation over three trials for the cats and dogs image classification experiment.}
    \vspace{1em}
    \begin{tabular}{@{}lcc@{}}
        \toprule
        \textbf{Method/Exp.} & \textbf{Para. Count} & \textbf{Test Accuracy} \\ 
        \midrule
        Baseline & 1.77M & $77.6\% \pm 0.86\%$ \\ 
        Unreg. N-ODE Trans. & 1.48M & $75.6\% \pm 0.48\%$ \\ 
        Reg. N-ODE Trans. & 1.48M & $75.6 \% \pm 0.03 \%$ \\ 
        Unreg. OT-Trans. & 1.48M & $78.2\% \pm 0.39\%$ \\ 
        OT-Trans. (Ours) & 1.48M & $\mathbf{79.0} \textbf{\%} \bm{\pm}
        \mathbf{0.31\%}$ \\ 
        \bottomrule
    \end{tabular}
    \label{tab:catdog}
\end{table}

\vspace{-1em}

\subsection{Sentiment Analysis}\label{subsec:imdb_exp}
We perform sentiment analysis on the IMDb movie review dataset~\cite{maas2011learning}, aiming to predict whether each movie review is positive or negative. We use an identical baseline transformer architecture as in~\cite{sander2022sinkformers}, which has six layers of transformer blocks. The OT-Transformer counterpart has only 3 layers, reducing the number of parameters of the transformer blocks by half.

We repeat the experiment for five random trials. In all trials, the unregularized N-ODE Transformer and OT-Transformer experienced issues with exploding gradients, resulting in NaN outputs. In order to estimate how the unregularized model would perform under more stable conditions, we impose a slight transport cost with $\lambda=0.01$. We note that the continuous-time models with slight and standard regularization completed all trials without issues. This shows the effectiveness of the transport cost regularization in stabilizing the training process and avoiding exploding gradients.

The best test accuracies are reported in~\Cref{tab:sentiment}. The baseline architecture achieved a test accuracy of 83.9\%. The N-ODE Transformers with slight and standard regularization report a test accuracy of 83.6\% and 83.9\%, respectively, which are not better than the baseline model. The N-ODE Transformer with slight regularization reports a test accuracy of 83.6\%. With a standard regularization, the test accuracy slightly increases to 83.9\%. However, both results are not better than that of the baseline model. The OT-Transformer with slight regularization reported a test accuracy of 82.7\%, which is subpar compared to the baseline model. 
On the other hand, the standard OT-Transformer achieves the best test accuracy of 84.6\%, which is 0.7\% higher than the baseline model, in spite of using a smaller model. The test accuracy is also 0.7\% higher than that of the N-ODE Transformer's. We note that with the incorporation of transport cost, the accuracy of N-ODE Transformer is improved by only 0.3\%. In contrast, the accuracy of OT-Transformer is boosted by 1.9\%. Again, this is likely due to that our continuous-time formulation is inherently more suited for transport cost regularization than that of N-ODE Transformer; see~\Cref{sec:related} for the more detailed discussion. 

\begin{figure}[h]
    \centering
    \includegraphics[width=0.48\textwidth]{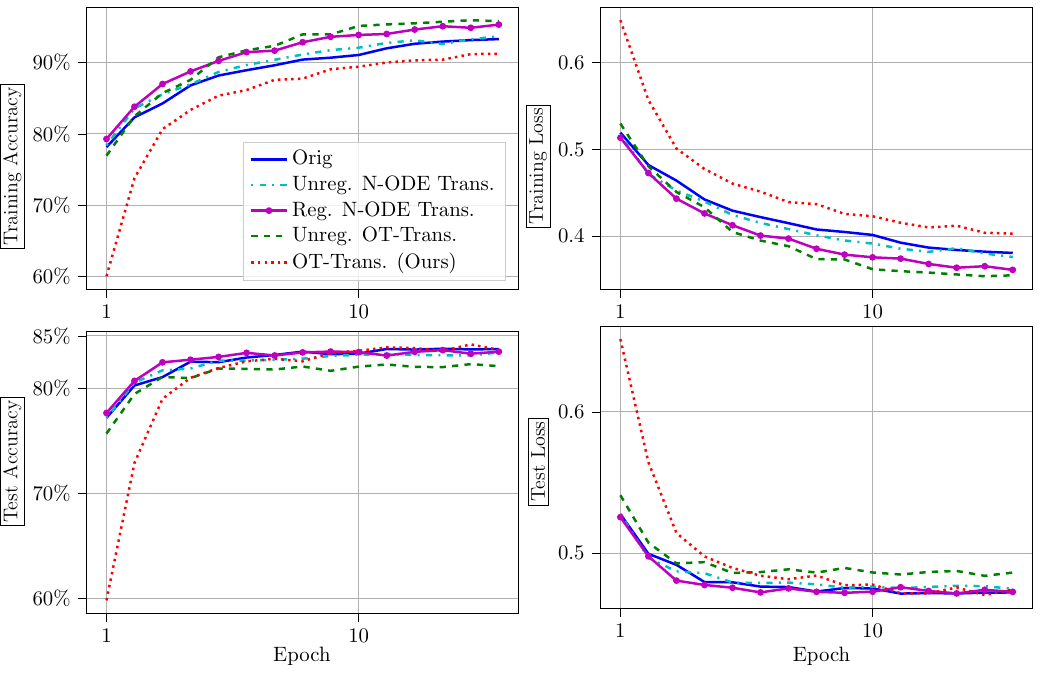}
    \vspace{-2em}
    \caption{Accuracy and data-fitting loss for the sentiment analysis experiment}
    \label{fig:sentiment}
\end{figure}

The learning curves are reported in~\Cref{fig:sentiment}. When we compare the results of the unregularized and regularized OT-Transformers, we see that the regularization effectively reduces overfitting by increasing training loss while simultaneously lowering test loss. Overall, we see that the combination of our continuous-in-time formulation and transport cost regularization enhances parameter efficiency and generalization of transformers.

\begin{table}[h]
    \centering
    \caption{Mean test accuracy and standard deviation over five trials for the sentiment analysis experiment. $^*$: The unregularized continuous-time models experienced gradient explosion. And we estimate their performance by using a slight regularization $\lambda=0.01$.}
    \vspace{1em}
    \begin{tabular}{@{}lcc@{}}
        \toprule
        \textbf{Method/Exp.} & \textbf{Para. Count} & \textbf{Test Accuracy} \\ 
        \midrule
        Baseline & 4.74M & $83.9\% \pm 0.26\%$ \\ 
        Unreg. N-ODE Trans. & 2.37M & $83.6\% \pm 0.40\%$$^*$ \\ 
        Reg. N-ODE Trans. & 2.37M & $83.9\% \pm 0.48\%$ \\ 
        Unreg. OT-Trans. & 2.37M & $82.7\% \pm 0.38\%$$^*$ \\ 
        OT-Trans. (Ours) & 2.37M & $\mathbf{84.6} \textbf{\%} \bm{\pm} \mathbf{0.55} \textbf{\%}$ \\ 
        \bottomrule
    \end{tabular}
    \label{tab:sentiment}
\end{table}

\vspace{-0.5em}

\section{Discussion and Summary}
We proposed OT-Transformer, a continuous-time formulation of transformers. OT-Transformer is flexible and is general-purpose, as it can be easily adapted to different variations of the vanilla transformer architecture, making it suitable for a wide class of tasks. It is also distinctive from existing continuous-time transformer architecture. Our training objective includes a transport cost regularization, which we justified through theory and extensive experimentation. In particular, we showed that the training problem is ill-posed without the regularization. We also illustrated that the regularization stabilizes the training process and enhances the generalization of our model. Through multiple tests across different applications, we demonstrate that our model improves the baseline transformer architecture in terms of parameter efficiency and accuracy, while reducing the variance among trials at the same time. This is particularly beneficial during inference; without the need for gradient tracking, our smaller models are more memory efficient. Contributing to the point, we also notice that it is possible to reduce the number of time steps at the cost of minor decrease in performance during inference, see \cref{sec:time_stepping_test}.
Most importantly, it outperforms the existing continuous-time transformer architecture. These results showcase the effectiveness and potential of our model.

\section*{Impact Statement}
This paper presents work whose goal is to advance the field of Machine Learning. There are many potential societal consequences of our work, none of which we feel must be specifically highlighted here.

\section*{Acknowledgements}
The authors would like to thank Lars Ruthotto and Tingwei Meng for the valuable advice and insightful discussion.

\bibliography{main}
\bibliographystyle{icml2024}

\newpage
\appendix
\onecolumn

\section{Different Time Stepping at Inference}
\label{sec:time_stepping_test}
In this section, we briefly discuss the numerical results of selecting different time step sizes at inference and the changes in model performance. It is evident in the Neural ODE literature, such as \cite{chen2018neural, onken2020discretize} that using smaller step sizes in time discretization improves integration accuracy and can enhance overall model performance. However, in \cite{onken2021ot} it is also pointed out that such requirement can be relaxed if the underlying dynamics is sufficiently regular, particularly at inference. We here use the aforementioned MNIST experiment in \cref{sec:experiment} to verify the performance change when using different time step sizes to evaluate a pretrained model. Note that the models reported in~\cref{sec:experiment} are trained using $20$ time steps from time $0$ to $T=1$. 

\begin{table}[h!]
\centering
\caption{Results for the MNIST example, tested using different number of time steps over trained models, here we use accuracy on the test data set to indicate performance.}
\begin{tabular}{|c|c|c|c|c|c|c|}
\hline
\diagbox{method}{number of time steps} & $1$ & $2$ & $4$ & $8$ & $16$ &$20$ \\ \hline
Unregularized OT-Transformer &  $18.2\%$            &  $42.0\%$     &    $85.1\%$  & $96.3\%$  & $96.7\%$  & $96.8\%$   \\ \hline
OT-Transformer &  $17.9\%$            &  $46.8\%$     &    $87.1\%$  & $96.6\%$  & $97.0\%$  & $97.1\%$   \\ \hline
\end{tabular}
\label{tab:step_size_mnist}
\end{table}

We display the results in \cref{tab:step_size_mnist}. Here, we test different number of time steps from $1$ to $20$ for both the unregularized model and the OT-Transformer model. Notice here first regularized models performance consistently better than that of the unregularized model, indicating the importance of OT regularization. More importantly we find that it is possible to reduce the number of time steps in evaluation with little decrease in model performance. Specifically we note that decreasing the number of time steps from $20$ to $8$ only resulted in about $0.5\%$ decrease in test accuracy. We believe this finding can be meaningful, as it suggests further efficiency improvement at the model deployment stage. However, additional testing may be required for other examples, we will leave further investigation of this point for future work.

\section{Experimental Details and Results}\label{sec:exp_details}
We report the detailed experimental setups here. We adapted the code provided by~\cite{sander2022sinkformers}, maintaining the same default data processing setup, hyperparameters, and other experimental settings as used in their implementation.

\paragraph{Point Cloud Classification}
We use the ModelNet40 dataset. For each instance, we uniformly sample 5000 points from each element in the dataset. We use a Set Transformer~\cite{lee2019set} with two Induced Self Attention Blocks (ISABs) in the encoder, where each ISAB contains two transformer blocks, and with a Pooling by Multihead Attention (PMA) Module in the decoder. We use an Adam optimizer, with batch size 64, 200 training epochs, and learning rate of $1 \times 10^{-3}$. For the baseline transformer model, the hidden dimensions of the ISABs are $d,k=256$, and for the continuous-time models, they are reduced to $200$. For the regularized N-ODE Transformer and OT-Transformer, the regularization hyperparameters are $\lambda=0.1$ and $\lambda=1$, respectively, as they provide the optimal performance in our tests. We use $T=1$ and a total of 8 time steps for the numerical integration.

\begin{table}[h]
    \centering
    \caption{Number of parameters for the transformer blocks, best and final test accuracies (with standard deviation) across five trials for the point cloud experiment.}
    \vspace{1em}
    \begin{tabular}{@{}lccc@{}}
        \toprule
        \textbf{Method/Exp.} & \textbf{Para. Count} & \textbf{Best Test Accuracy} & \textbf{Final Test Accuracy}\\ 
        \midrule
        Baseline & 0.86M & $87.4\% \pm 0.45\%$ & $86.6\% \pm 0.67\%$ \\ 
        Reg. N-ODE Trans. & 0.65M & $87.5\% \pm 0.51\%$ & $86.7\% \pm 0.43\%$  \\ 
        OT-Trans. (Ours) & 0.65M & $\mathbf{89.9} \textbf{\%} \bm{\pm}\mathbf{ 0.42} \textbf{\%}$ & $\mathbf{89.3} \textbf{\%} \bm{\pm}\mathbf{ 0.69} \textbf{\%}$ \\ 
        \bottomrule
    \end{tabular}
\end{table}

\paragraph{MNIST Classification}
We use a Vision Transformer (ViT)~\cite{dosovitskiy2021an} with self-attention layer with a single head. The patch size is $7 \times 7$. We use an Adam optimizer. The number of epochs is 45 and the batch size is 100. The learning rate is set to $5\times 10^{-4}$ for the first 35 epochs, then decreased to $5\times 10^{-5}$ until the 41st epoch, at which point it is reduced to $5\times 10^{-6}$. For the baseline model, the hidden dimensions $d$ and $k$ are 128. For the continuous-time models, they are reduced to 64. For OT-Transformer, the regularization hyperparameter is $\lambda=0.01$ as it provides the optimal performance in our tests. We use $T=1$ and a total of 20 time steps for the numerical integration.

\begin{table}[h]
    \centering
    \caption{Number of parameters for the transformer blocks, best and final test accuracies (with standard deviation) across five trials for the MNIST image classification experiment.}
    \vspace{1em}
    \begin{tabular}{@{}lccc@{}}
        \toprule
        \textbf{Method/Exp.} & \textbf{Para. Count} & \textbf{Best Test Accuracy} & \textbf{Final Test Accuracy} \\ 
        \midrule
        Baseline & 93k & $93.0\% \pm 0.69\%$ & $93.0\% \pm 0.67\%$ \\ 
        Unreg. OT-Trans. & 18k & $96.8\% \pm 0.23\%$ & $96.8\% \pm 0.25\%$ \\ 
        OT-Trans. (Ours) & 18k & $\mathbf{97.1} \textbf{\%} \bm{\pm}
        \mathbf{0.16\%}$ & $\mathbf{97.1} \textbf{\%} \bm{\pm}
        \mathbf{0.15\%}$ \\ 
        \bottomrule
    \end{tabular}
\end{table}

\paragraph{Cats and Dogs Classification}
We again use ViT. The patch size is $16 \times 16$. We use an Adam optimizer. The learning rate is $3 \times 10^{-5}$. The number of epochs is 250, and the batch size is 64. The hidden dimensions $d$ and $k$ are 128. For the baseline model, it has 6 transformer blocks. For the continuous-time models, the number of transformer blocks is reduced to 5. For the regularized N-ODE Transformer and OT-Transformer, the regularization hyperparameters are $\lambda=0.005$ and $\lambda=0.01$, respectively, as they provide the optimal performance in our tests. We use $T=1$ and a total of 20 time steps for the numerical integration.

\begin{table}[h]
    \centering
    \caption{Number of parameters for the transformer blocks, best and final test accuracies (with standard deviation) across three trials for the cats and dogs image classification experiment.}
    \vspace{1em}
    \begin{tabular}{@{}lccc@{}}
        \toprule
        \textbf{Method/Exp.} & \textbf{Para. Count} & \textbf{Best Test Accuracy} & \textbf{Final Test Accuracy} \\ 
        \midrule
        Baseline & 1.77M & $79.3\% \pm 0.52\%$ & $77.6\% \pm 0.86\%$ \\ 
        Unreg. N-ODE Trans. & 1.48M & $76.4\% \pm 0.37\%$ & $75.6\% \pm 0.48\%$ \\ 
        Reg. N-ODE Trans. & 1.48M & $76.4\% \pm 0.30\%$ & $75.6 \% \pm 0.03 \%$ \\ 
        Unreg. OT-Trans. & 1.48M & $78.8\% \pm 0.63\%$ & $78.2\% \pm 0.39\%$ \\ 
        OT-Trans. (Ours) & 1.48M & $\mathbf{79.5} \textbf{\%} \bm{\pm}
        \mathbf{0.46\%}$ & $\mathbf{79.0} \textbf{\%} \bm{\pm}
        \mathbf{0.31\%}$ \\ 
        \bottomrule
    \end{tabular}
\end{table}

\paragraph{Sentiment Analysis}
We follow~\cite{sander2022sinkformers} to use a baseline model with 6 layers of transformer blocks. For the continuous-time models, the number of transformer blocks is reduced to 3. We use an Adam optimizer with 15 epochs. The learning rate is $1 \times 10^{-4}$ for the first 12 epochs and $1 \times 10^{-5}$ afterward. The batch size is 64. The hidden dimensions $d$ and $k$ are 256. The batch size is 64. 
For both the regularized N-ODE Transformer and OT-Transformer, the regularization hyperparameter is $\lambda=0.5$, as it provides the optimal performance in our tests.
We use $T=1$ and a total of 8 time steps for the numerical integration.

\begin{table}[h]
    \centering
    \caption{Number of parameters for the transformer blocks, best and final test accuracies (with standard deviation) across five trials for for the sentiment analysis experiment. $^*$: The unregularized continuous-time models experienced gradient explosion. And we estimate their performance by using a slight regularization $\lambda=0.01$.}
    \vspace{1em}
    \begin{tabular}{@{}lccc@{}}
        \toprule
        \textbf{Method/Exp.} & \textbf{Para. Count} & \textbf{Best Test Accuracy} & \textbf{Final Test Accuracy} \\ 
        \midrule
        Baseline & 4.74M & $83.9\% \pm 0.26\%$ & $\mathbf{83.7} \textbf{\%} \bm{\pm} \mathbf{0.21} \textbf{\%}$ \\ 
        Unreg. N-ODE Trans. & 2.37M & $83.6\% \pm 0.40\%$$^*$ & $83.4\% \pm 0.40\%$$^*$  \\ 
        Reg. N-ODE Trans. & 2.37M & $83.9\% \pm 0.48\%$ & $83.5\% \pm 0.84\%$ \\ 
        Unreg. OT-Trans. & 2.37M & $82.7\% \pm 0.38\%$$^*$ & $82.1\% \pm 0.89\%$$^*$\\ 
        OT-Trans. (Ours) & 2.37M & $\mathbf{84.6} \textbf{\%} \bm{\pm} \mathbf{0.55} \textbf{\%}$ & $83.7\% \pm 0.86\%$ \\ 
        \bottomrule
    \end{tabular}
\end{table}

\end{document}

%% file: abbrv.tex
\newcommand{\bfK}{{\bf K}}

\newcommand{\bfP}{{\bf P}}
\newcommand{\bfQ}{{\bf Q}}

\newcommand{\bfT}{{\bf T}}

\newcommand{\bfV}{{\bf V}}
\newcommand{\bfW}{{\bf W}}
\newcommand{\bfX}{{\bf X}}

\newcommand{\bfZ}{{\bf Z}}

\newcommand{\bfx}{{\bf x}}
\newcommand{\bfy}{ {\bf y}}
\newcommand{\bfu}{{\bf u}}

\newcommand{\bfz}{{\bf z}}

\newcommand{\bftheta}{{\boldsymbol \theta}}

\newcommand{\bfgamma}{{\boldsymbol \gamma}}